**¿Confiar en ChatGPT? Cuando una variación sutil en el prompt puede modificar significativamente los resultados.**

**Autores**

Cuellar, Jaime E.
jaimecuellar@javeriana.edu.co
Facultad de Comunicación y Lenguaje
Pontificia Universidad Javeriana (Colombia)
https://orcid.org/0009-0004-0858-2823

Moreno-Martínez, Óscar. Ph.D.
morenoo@javeriana.edu.co
Facultad de Comunicación y Lenguaje
Pontificia Universidad Javeriana (Colombia)
https://orcid.org/0000-0001-9735-9455

Torres-Rodríguez, Paula Sofía
torres.ps@javeriana.edu.co
Facultad de Ciencias
Pontificia Universidad Javeriana (Colombia)
https://orcid.org/0009-0008-8108-9153

Pavlich-Mariscal, Jaime Andres. Ph.D.
jpavlich@javeriana.edu.co
Facultad de Ingeniería
Pontificia Universidad Javeriana (Colombia)
https://orcid.org/0000-0002-3892-6680

Micán-Castiblanco, Andrés Felipe
a.mican@javeriana.edu.co
Facultad de Comunicación y Lenguaje
Pontificia Universidad Javeriana (Colombia)
https://orcid.org/0000-0002-0049-750X

Torres-Hurtado, Juan Guillermo. Ph.D.
juangtorresh@javeriana.edu.co
Facultad de Ingeniería
Pontificia Universidad Javeriana (Colombia)
https://orcid.org/0000-0001-8912-9289

**Resumen**

En la actualidad una pregunta fundamental para las ciencias sociales es ¿qué tanto podemos confiar en modelos predictivos de alta complejidad como ChatGPT? Para acercarse a esta pregunta este estudio testea la hipótesis de que cambios sutiles en la estructuración de *prompts* no producen variaciones significativas en los resultados de clasificación del análisis de sentimientos de polaridad producido por el *Large Language Models* GPT-4o mini. Utilizando una base de datos de 100.000 comentarios en español sobre cuatro presidentes latinoamericanos, el modelo clasificó los textos como positivos, negativos o neutrales en 10 ocasiones, variando sutilmente los prompts empleados. La metodología experimental incluyó análisis exploratorios y confirmatorios para identificar discrepancias significativas entre las clasificaciones.

Los resultados revelan que incluso modificaciones menores en los *prompts*, como cambios léxicos, sintácticos o modales, e incluso su desestructuración, afectan la clasificación. En ciertos casos, el modelo produjo respuestas inconsistentes, como mezclar categorías, ofrecer explicaciones no solicitadas o usar idiomas distintos al español. El análisis estadístico, mediante Chi-cuadrado, confirmó diferencias significativas en la mayoría de las comparaciones entre prompts, salvo en un caso con estructuras lingüísticas altamente similares.

Estos hallazgos cuestionan la robustez y confiabilidad de los *Large Language Models* para tareas de clasificación, al tiempo que resaltan su vulnerabilidad a variaciones en las instrucciones. Además, se evidenció que la falta de gramática estructurada en los prompts incrementa la frecuencia de alucinaciones. La discusión enfatiza que la confianza en *Large Language Models* no solo se basa en su desempeño técnico, sino también en la relación social e institucional que sustenta su uso.

**Palabras clave**

*Large Language Models (LLM)*, ChatGPT, confianza, robustez, español, análisis sentimientos

**Introducción**

En la actualidad, las herramientas del *Natural Language Processing* (NLP) son ampliamente usadas en diversos contextos académicos y empresariales. Estas herramientas sirven para analizar, interpretar o generar lenguaje natural ya sea escrito o hablado (Elliott 2022). Las ciencias sociales las han empleado para, por ejemplo, analizar procesos electorales, luchas sociales o sesgos políticos en redes sociales (Saucedo 2018; Khatua y Nejdl 2021; Olabanjo et al. 2023;Fundación Karisma 2023). Estas herramientas, igualmente, han estado en el centro del debate académico en campos como la historia, la archivística y el patrimonio debido a los acelerados procesos de digitalización de archivos, transcripción automática de fuentes orales o gestión/conservación cultural (Sporleder 2010; Pessanha y Salah 2021; Gray et al. 2022; Marxen 2023). En contextos empresariales, se han utilizado para analizar reseñas de productos y servicios, realizar marketing y publicidad en redes sociales o desarrollar *chatbots* (Ermakova et al. 2021; Kusal et al. 2021).

Los avances más recientes en el campo del NLP son los *Large Language Models* (LLMs), definidos como modelos que cuentan con un gran número de parámetros y capas, y que han sido entrenados utilizando vastos conjuntos de textos de diversas fuentes (Brown et al. 2020). Tras recibir una instrucción en forma de *prompt*, los LLMs pueden generar respuestas en texto que imitan verosímilmente el estilo y el contenido que un humano produciría. Los LLMs más avanzados son accesibles por Internet a través de una *Application Programming Interface* (API), interfaces de comunicación con una sintaxis estandarizada que definen cómo dos o más componentes de software interoperan (Jacobson et al. 2012; Mitchell 2018; Jünger 2021).[1] ChatGPT se ha consolidado como uno de los LLMs más populares. Apenas dos meses después de su lanzamiento, a finales de 2022, alcanzó más de 100 millones de usuarios activos mensuales (Data.ai 2023).

Como propone Zhang et al. (2023), los LLMs podrían producir resultados disímiles incluso cuando las indicaciones son semánticamente similares. Estas diferencias en los resultados actualizan una discusión que ya ha sido aproximada por los Estudios Sociales de Ciencia y Tecnología (STS, por sus siglas en inglés): ¿qué tanto podemos confiar en modelos predictivos de alta complejidad como ChatGPT? Aparentemente, esta confianza debería basarse en la robustez matemática y estadística del modelo. Respondiendo a esta pregunta, este artículo pone a prueba la hipótesis de que cambios sutiles en la estructuración de *prompts* no producen variaciones significativas en los resultados de clasificación del análisis de sentimientos de polaridad producido por el LLM GPT-4o mini. Para esto, se puso a prueba la robustez del modelo, pidiéndole que clasificara una base de datos de 100.000 comentarios de Youtube a partir de diez prompts diferentes para evaluar si existen diferencias significativas entre sus respuestas.

---

[1] Entendemos a las APIs y los LLMs como partes de un mismo todo, como una unidad. Las APIs son las interfaces para conectar con el modelo, son las que definen términos de operación y posibilidades de función (Weltevrede 2016). En este sentido, existe una interdependencia entre API y LLM separable solo para efectos analíticos (Nogara et al. 2023). OpenAI, la empresa detrás de ChatGPT, autoriza el uso de sus LLM a partir de pagos.

El artículo se divide en cuatro secciones. La primera, la revisión de literatura, repasa investigaciones y marcos conceptuales sobre los LLMs, el análisis de sentimientos, la construcción de prompts, la confianza y la robustez. La segunda, la metodología, describe el proceso de recolección, estructuración, procesamiento y análisis de los comentarios, así como el desarrollo de los diez *prompts* construidos, y los métodos exploratorios y confirmatorios que se usaron para evaluar la robustez del modelo. La tercera, los resultados, presenta los hallazgos de lo anteriormente descrito y los interpreta para identificar si existen o no diferencias significativas. Y la cuarta, la discusión, problematiza las implicaciones de confiar en LLM con robustez inconsistente, especialmente en un contexto caracterizado por su adopción masiva y su creciente importancia como soportes de procesos de investigación y de toma de decisiones.

**Revisión de literatura**

**LLM**

Como se describió anteriormente, los LLMs son modelos basados en redes neuronales (Brown et al. 2020) que reciben instrucciones en forma de *prompt* y pueden generar respuestas en texto. Su impacto en diferentes áreas del conocimiento ha sido inmediato. En la medicina, por ejemplo, se han usado para proporcionar información de salud personalizada a los pacientes y para brindar servicio al cliente 24 horas a través de *chatbots* (Biswas 2023); en la religión, ha apoyado la exégesis del Corán y del pensamiento islámico moderno (El Ganadi et al. 2023); en la economía, ha asistido el procesamiento de indicadores financieros (Li et al. 2023) y, en la comunicación, ha servido para identificar información falsa (Chen et al. 2022) y para analizar comentarios de redes sociales chinas con respecto a, por ejemplo, la guerra entre Ucrania y Rusia (Rogers y Zhang 2024).

A pesar del reciente auge de los LLM como ChatGPT, ya se han documentado varias limitaciones del modelo. Borji (2023) construye una tipología de once fallas de ChatGPT 3.5:

equivocaciones de razonamiento lógico, errores factuales, inexactitudes matemáticas, dificultades para generar fragmentos de código funcionales, sesgos que reproducen estereotipos sociales, errores lingüísticos, expresiones artificiales o fingidas de emociones, incapacidad para conectar con las personas a través de contenidos emocionalmente resonantes, impericia para ofrecer perspectivas originales, falta de claridad en cómo toma las decisiones y, por último, difusión de información errónea o propagandística.

Estas fallas son difíciles de explicar ya que estos LLMs funcionan como cajas negras (Yadav y Vishwakarma 2020; Ermakova et al. 2021; Latour 2021). Es decir, son modelos cuyos datos de entrada, resultados e incluso posibles limitaciones son conocidos, pero cuyo proceso interno continua siendo opaco. La manera en que realizan sus predicciones o análisis se basa en un conjunto de operaciones complejas, de las que solo puede observarse la superficie (Ermakova et al. 2021). No solo los modelos analizados son cajas negras, sino que sus APIs patentadas también lo serían, ya que su funcionamiento está escasamente documentado (Nogara et al. 2023).

**Análisis de sentimientos**

Entre las múltiples tareas que realizan los LLM se destaca la de clasificación de datos. El proceso de clasificación permite distribuir datos a partir de clases previamente definidas. Un ejemplo de esto es el análisis de sentimientos que es un proceso que permite clasificar texto en categorías relevantes para dar cuenta de las opiniones inmersas en este (Ermakova et al. 2021), utilizando herramientas computacionales y criterios preestablecidos o aprendidos. Dentro del análisis de sentimientos, el análisis de polaridad que clasifica los textos en tres categorías: positivo, negativo y neutral (Kusal et al. 2021 ; Ermakova et al. 2021). Su objetivo es asignar cada elemento de un conjunto de textos (tweets, comentarios de YouTube o cualquier otro tipo) a la categoría que mejor se ajuste a su contenido.

El análisis de sentimientos ha sido ampliamente utilizado para investigar fenómenos sociales en diversos contextos. Por ejemplo, Chandra y Ritij (2021) aplicaron esta técnica para predecir los resultados de las elecciones presidenciales entre Joe Biden y Donald Trump en el 2020. De manera similar, Li et al. (2024) emplearon el análisis de sentimientos como base para desarrollar una métrica del impacto de los rumores en redes sociales en términos de daño. En el ámbito de la salud, Braig et al. (2023) exploraron cómo el análisis de sentimientos en Twitter puede ofrecer información clave para gestionar la pandemia del COVID-19. En el campo educativo, Sultana et al. (2018) utilizaron el análisis de sentimientos en datos académicos para predecir el rendimiento estudiantil. Por último, Loomba et al. (2024) aplicaron esta técnica a datos de Twitter para analizar las percepciones de la comunidad india sobre las criptomonedas.

**Construcción de *prompts***

El *prompt* es la instrucción escrita que se le da a un LLM para que realice una tarea o consulta específica guiando su comportamiento y generando los resultados deseados. En este caso, la tarea consistió en clasificar los comentarios de vídeos de YouTube en positivos, negativos o neutrales. Dado que es posible solicitar la misma tarea con múltiples instrucciones, es indispensable volver la atención sobre la forma en que se estructura el *prompt* para comprender cuál es la mejor manera de comunicarse con los LLM.

La ingeniería de instrucciones o *prompt engineering* es una disciplina relativamente reciente que se refiere a la práctica de desarrollar y optimizar *prompts* para utilizar de manera efectiva LLMs, en particular, en tareas de procesamiento de lenguaje natural de diferentes áreas (Giray 2023). De acuerdo con la *Prompt Engineering Guide* (DAIR.AI 2023), desarrollada por el grupo de *Democratizing Artificial Intelligence Research, Education, and Technologies*, un *prompt* puede contener cualquiera de los siguientes elementos: una instrucción o tarea específica que se desea que el modelo realice, un contexto o información adicional que puede orientar al modelo hacia respuestas más precisas, un dato de entrada o pregunta que se busca resolver y un indicador de salida que se refiere al tipo o formato esperado del resultado.

En los últimos años, se han elaborado guías (Ozdemir 2023; Marvin et al. 2023) y recomendaciones para la construcción de *prompts* entre las que se incluye cuidar su especificidad y su claridad, estructurar el tipo de input, especificar el formato del resultado deseado, utilizar delimitadores, descomponer oraciones complejas, entre otras. De manera complementaria, la ingeniería avanzada de *prompts* sugiere diferentes técnicas para redactarlos entre las que se puede mencionar: *zero-shot prompting* (Wei et al. 2021), *few-shot prompting* (Reynolds y McDonell 2021), *chain-of-thought prompting* (Wei *et al*. 2022), *meta prompting* (Zhang et al. 2023), *self-consistency* (Wang et al. 2022), *generate knowledge prompting* (Liu et al. 2021), *tree of thoughts* (Long 2023), etc. En síntesis, con los distintos desarrollos de la ingeniería de *prompts,* se han ido sofisticando sus técnicas.

**Confianza y robustez**

La confianza en los LLM se ha convertido en un tema fundamental, especialmente a medida que estos modelos se integran en las actividades cotidianas y profesionales de las personas. Un interrogante clave en este contexto es: ¿es posible confiar en estos modelos? Este es un debate complejo y desafiante, que abarca tanto cuestiones técnicas como éticas y que aún está lejos de tener una respuesta definitiva. Para las ciencias sociales, es esencial no solo abordar este tema desde una perspectiva crítica y teórica, sino también emprender aproximaciones empíricas que permitan aportar evidencia y perspectivas concretas. En este sentido, Zanotti et al. (2024) sugiere que la confianza no es solo un resultado deseable en el uso de los LLM, sino una condición necesaria para la adopción y la aplicación efectiva de estos modelos en diversos ámbitos.

Por un lado, diversos autores, desde la ingeniería, han explorado las condiciones para que los LLM sean confiables. Bolton et al. (2024), por ejemplo, sugieren que la robustez en el procesamiento de variaciones de prompts, la alta capacidad de recordar información relevante y la ausencia de alucinaciones son criterios fundamentales para el uso confiable de estos modelos. Por su parte, Huang et al. (2024) plantean una serie de dimensiones necesarias para construir confianza en los LLM, que incluyen la veracidad, la seguridad, la equidad, la robustez, la privacidad, la ética de las máquinas, la responsabilidad y la conciencia. Asimismo, Majeed y Hwang (2024) destacan la importancia de incluir información transparente sobre las fuentes de datos empleadas, la provisión de múltiples respuestas para una misma consulta y la evaluación de la credibilidad de las fuentes usadas en el entrenamiento del modelo. Por lo tanto, desde esta perspectiva se entiende que para tener confianza en los LLM es necesario el cumplimiento de métricas y condiciones preestablecidas.

Siguiendo con la idea anterior, uno de los elementos centrales que la literatura identifica como esencial para generar confianza en los LLM es la robustez. Bolton et al. (2024) definen la robustez como la capacidad de los LLM para resistir variaciones no semánticas en los *prompts*, es decir, deben ser robustos ante modificaciones en el *prompt* que no alteren sustancialmente su significado. Este aspecto es vital, ya que permite que los LLM mantengan consistencia en sus respuestas aun cuando los inputs presentan ligeras variaciones. A su vez, Huang et al. (2024) amplían esta idea al definir la robustez como la habilidad del modelo para mantener su rendimiento en diversas circunstancias, incluyendo cambios en las entradas, variaciones contextuales o la presencia de ruido o de errores en los datos. Koubaa et al. (2023) entienden este concepto como la habilidad para mantener un rendimiento constante ante entradas o perturbaciones inesperadas, asegurando la fiabilidad y coherencia de las predicciones en distintos escenarios. En el caso de un chatbot utilizado para generar respuestas, esta cualidad se evidencia en su capacidad para resistir una variedad de estilos de lenguaje y de temas, garantizando así una experiencia consistente para el usuario.

Por otro lado, desde las ciencias sociales, para Cook y Santana (2020) la confianza es un atributo relacional que emerge entre actores y no tanto un atributo individual o autocontenido. Según Cook, Hardin y Levi (2005), la confianza emerge a través de intereses encapsulados, es decir, cuando una de las partes percibe que la otra tiene incentivos para actuar en su beneficio, motivada por su compromiso con la relación y su interés en mantener una reputación de confiabilidad. De todas formas, confiar en LLM agregaría una capa de complejidad. Autores como Taddeo (2009) y Grodzinsky et al. (2010) hablan de *e-trust* para estudiar la relación de confianza entre humanos y agentes artificiales. Para Grodzinsky et al. (2020), un agente artificial es una máquina creada por personas para funcionar sin necesidad de intervención humana constante. La particularidad de estos agentes radica en que pueden modificar su estado interno dependiendo de su interacción con el entorno a lo largo del tiempo. Estas variaciones podrían socavar la confianza en ellos.

En otras palabras, confiar en agentes artificiales es diferente a confiar en otros artefactos (como una balsa o una bicicleta). Los LLM, en tanto agentes artificiales, no requieren por largas temporalidades de la intervención humana para funcionar. Además, pueden entrenarse y eventualmente presentar variaciones en sus respuestas a una misma tarea. Taddeo (2009) y Grodzinsky et al. (2020) concuerdan en que la opacidad, las actualizaciones de software y su comportamiento incierto, derivado de su aprendizaje, presentan desafíos únicos para la confianza en este tipo de agentes.

Sociólogos de la ciencia como Shapin (1995) reivindican la definición relacional de la confianza. En su caso de estudio sobre el conocimiento científico de la Inglaterra del siglo XVII, Shapin nota que la producción de dicho conocimiento estuvo más vinculada a relaciones de confianza entre humanos que a la noción de "verdad científica" de la medicina. Para Shapin, la palabra "confianza" implica una relación, mientras que la "verdad" apunta a una esencia; por ello, sostiene que es más adecuado hablar de confianza que de verdad en este contexto. Los LLM, en este sentido, no mienten ni dicen la verdad; simplemente se trata de si confiar o no en ellos. Rolin (2020) amplía esta concepción al señalar que la confianza epistémica, es decir, la confianza en el conocimiento científico, no solo se orienta hacia científicos individuales o equipos de investigación, sino también hacia prácticas sociales e instituciones científicas.

Además, los STS han desarrollado marcos útiles para ampliar la discusión sobre la confianza en modelos tecnológicos complejos. Nowotny (2021) sostiene que la confianza en la inteligencia artificial no depende únicamente de su robustez técnica, sino de factores tales como la explicabilidad, la transparencia y la adaptabilidad, entre otros. El atractivo de los algoritmos predictivos y sus respuestas aparentemente precisas radica en que generan una falsa sensación de certeza y de control, lo que, a su vez, fomenta una confianza ciega, e incluso una dependencia. En otras palabras, Nowotny (2021) interpreta la confianza en la robustez de un modelo como una manifestación del deseo humano de anular la incertidumbre y recuperar el control sobre el futuro. El ser humano persigue esta anhelada certeza a través de las operaciones predictivas algorítmicas, lo que constituye una ilusión de seguridad.

**Metodología**

El artículo sigue una metodología cuantitativa experimental (Creswell y Creswell 2018) que consiste en manipular una o más variables para evaluar cómo estos cambios impactan en el resultado. Específicamente, se analiza si ajustes sutiles en la estructuración de *prompts* producen variaciones significativas en el análisis de sentimientos de polaridad que realiza el LLM GPT-4o mini. Cuatro fases metodológicas se llevaron a cabo: 1) recolección y estructuración de datos 2) diseño de *prompts* 3) procesamiento de datos 4) análisis cuantitativo.[2]

**Recolección y estructuración**

Para construir el corpus, se descargaron 100.000 comentarios en español sobre cuatro presidentes latinoamericanos: Andrés Manuel López Obrador de México, Nayib Bukele de El Salvador, Javier Milei de Argentina y Gustavo Petro de Colombia[3] a través de la API de YouTube. Estos comentarios fueron publicados en 24 canales periodísticos locales e internacionales entre enero y junio de 2024[4]. La búsqueda de los videos dentro de los canales seleccionados se generó a partir de palabras clave organizadas de la siguiente manera: "(País) + (nombre del Presidente) + 'presidente' ". Esta conjugación de palabras permitió focalizar videos y comentarios específicamente sobre los cuatro líderes políticos.

---

[2] El código utilizado para el procesamiento de datos y análisis puede consultarse en el siguiente repositorio de GitHub: https://github.com/sofi-devv/repositorioInvestigacion.
[3] Reconocemos que Andrés Manuel López Obrador dejó de ser presidente de México el 30 de septiembre de 2024. Al momento de descargar la información aún era presidente en ejercicio.
[4] La información completa se encuentra en el Anexo 1. Cuando decimos canales periodísticos locales nos referimos a medios del país donde el presidente es gobernante.

La base de datos se construyó con la intención de testear el resultado del proceso de clasificación del análisis de sentimientos en español, para lo cual fue indispensable buscar alta polarización en los comentarios descargados. Al recolectar comentarios sobre presidentes en ejercicio, especialmente acerca de presidentes percibidos como de izquierda o de derecha en el espectro ideológico, se pretendió recoger controversias manifiestas, tanto positivas, como negativas y neutrales. Además, al tener medios locales e internacionales de cuatro países latinoamericanos, se optó por nutrir la base de datos con distintos dialectos de español.

**Diseño de prompts**

Para este estudio se utilizaron diez prompts construidos con la técnica de *zero-shot prompting* (Tabla 1), dado que se buscaba que el modelo realizara una tarea sobre datos nuevos sin necesidad de ningún entrenamiento adicional. Se emplearon dos prompts-base o semilla más ocho variaciones en términos lingüísticos introducidas por el equipo investigador. En el grupo A, se incluye el *prompt*-base 1 sugerido por OpenAI (2023d) para realizar análisis de sentimientos y sus 4 variaciones (*prompts* 3, 5, 7 y 9). En el grupo B, se presenta el *prompt*-base o 2, basado en el usado por Zhang et al. (2023), para realizar la misma tarea y 4 variaciones (*prompts* 4, 6, 8 y 10). Los cambios en la estructuración lingüística de los *prompts*, se describen a continuación.

**Tabla 1: prompts utilizados para el análisis**

| Grupo | Tipo de prompt | Identificador | Prompt |
|---|---|---|---|
| Grupo A (Derivado prompt sugerido por Open AI) | Prompt de referencia | 1 | Como IA con experiencia en análisis de lenguaje y emociones, su tarea es analizar el sentimiento del siguiente texto. Considere el tono general de la discusión, la emoción transmitida por el lenguaje utilizado y el contexto en el que se utilizan las palabras y frases. Indique si el sentimiento es generalmente positivo, negativo o neutral y proporcione una etiqueta sin ningún texto adicional. |
| | Cambio sintáctico | 3 | Indique si el sentimiento es generalmente positivo, negativo o neutral y proporcione una etiqueta sin ningún texto adicional. Considere el tono general de la discusión, la emoción transmitida por el lenguaje utilizado y el contexto en el que se utilizan las palabras y frases. Como IA con experiencia en análisis de lenguaje y emociones, su tarea es analizar el sentimiento del siguiente texto. |
| | Cambio léxico-semántico | 5 | Como IA experto en procesamiento de lenguaje natural y análisis de sentimientos, su tarea es analizar el sentimiento del siguiente texto. Considere el tono general del comentario, los sentimientos transmitidos por el texto utilizado y el contexto en el que se utilizan las palabras y oraciones. Indique si el sentimiento es generalmente positivo, negativo o neutral y proporcione una etiqueta sin ningún texto adicional. |
| | Cambio modal-pragmático | 7 | Como IA con experiencia en análisis de lenguaje y emociones, debes analizar el sentimiento del siguiente texto. Puedes considerar el tono general de la discusión, la emoción transmitida por el lenguaje utilizado y el contexto en el que se utilizan las palabras y frases. Tienes que indicar si el sentimiento es generalmente positivo, negativo o neutral y proporcione una etiqueta sin ningún texto adicional. |
| | Prompt desestructurado | 9 | IA experiencia análisis lenguaje emociones tarea analizar sentimiento siguiente texto Considere tono general discusión emoción transmitida lenguaje utilizado contexto utilizan palabras frases indique sentimiento generalmente positivo negativo neutral proporcione etiqueta ningún texto adicional |
| Grupo B (Derivado prompt sugerido por la literatura) | Prompt de referencia | 2 | Realice la tarea de clasificación de sentimientos. Dado el texto, asigne una etiqueta de sentimiento entre positivo, negativo o neutral. Devuelva únicamente la etiqueta sin ningún otro texto. |
| | Cambio sintáctico | 4 | Devuelva únicamente la etiqueta sin ningún otro texto. Dado el texto, asigne una etiqueta de sentimiento entre positivo, negativo o neutral. Realice la tarea de clasificación de sentimientos. |
| | Cambio léxico-semántico | 6 | Realice la tarea de clasificación de sentimientos. Dado el comentario, asigne una etiqueta de sentimiento entre positivo, negativo o neutral. Devuelva únicamente la etiqueta sin ningún otro texto. |
| | Cambio modal-pragmático | 8 | Debes realizar la tarea de clasificación de sentimientos. Dado el texto, puedes asignar una etiqueta de sentimiento entre positivo, negativo o neutral. Tienes que devolver únicamente la etiqueta sin ningún otro texto. |
| | Prompt desestructurado | 10 | Realice tarea clasificación sentimientos dado texto asigne etiqueta sentimiento entre positivo negativo neutral devuelva únicamente etiqueta |

**Fuente:** Elaboración propia

El *prompt* de referencia en el análisis de sentimientos establece una instrucción general para que el modelo examine el sentimiento expresado en el texto y devuelva una etiqueta correspondiente, como positivo, negativo o neutral. En el *prompt* oficial sugerido por OpenAI, número 1, se genera un contexto, se refuerza la claridad de esta tarea y se incluye un recordatorio de no devolver ningún otro texto. En contraste, el *prompt* basado en la literatura, número 2, simplifica aún más las instrucciones (Krause y Vossen 2024), orientando al modelo a devolver solo la etiqueta de sentimiento, sin agregar contexto adicional.

Para los prompts 3 y 4 se realizó un cambio sintáctico de movimiento interoracional sin variación en el significado central. Es decir, se cambió el orden de las oraciones de cada *prompt*. Tal modificación buscó guiar al modelo para que devolviera exclusivamente la etiqueta y situó la instrucción en la cabeza de la oración, minimizando ambigüedades potenciales en las instrucciones. Una vez más se enfatizó que el resultado deseado debía limitarse a una sola palabra, como positivo, neutral o negativo, dado el tipo de análisis de sentimiento de la tarea. El número 3 hace una cambio sintáctico de variación intraoracional del prompt 1, mientras que el número 4 lo hace del prompt 2.

En los *prompts* 5 y 6 se realizó una variación léxico-semántica. En lugar de utilizar las palabras 'experticia en lenguaje' y 'análisis de emociones', se utilizaron las unidades 'procesamiento de lenguaje natural' y 'análisis de sentimientos' con la intención de probar la robustez del modelo con relación al reconocimiento y al procesamiento de términos sinónimos o relacionados semánticamente al momento de clasificar un sentimiento como positivo, negativo o neutral. El número 5 hace una variación léxico-semántica del prompt 1, mientras que el número 6 lo hace del prompt 2.

En los *prompts* 7 y 8 se introdujo un cambio modal-pragmático. Las variaciones en los prompts que incluyen modales, como "debe," "puede" o "necesita", exploran el efecto que tiene el nivel de obligación o énfasis en la instrucción sobre la respuesta del modelo. En particular, en términos lingüísticos y de la interfaz semántico-pragmática, los verbos modales imponen distintos grados de fuerza directiva, lo cual puede influir en la manera en que el modelo ejecuta la tarea. Así, un prompt que utiliza "debe" implica una mayor obligación y podría llevar al modelo a responder con mayor exactitud a la instrucción, mientras que un modal como "puede" introduce cierta flexibilidad epistémica, lo que puede suponer mayor variabilidad en las respuestas. El número 7 hace una variación modal del prompt 1, mientras que el número 8 lo hace del prompt 2.

Por último, los *prompts* 9 y 10 fueron diseñados sin palabras gramaticales ni puntuación para evaluar cómo el modelo maneja instrucciones cargadas semánticamente, pero fragmentadas gramaticalmente. En este sentido, se buscó poner a prueba la capacidad del modelo para interpretar correctamente el mensaje subyacente, con formas lingüísticas desestructuradas. Un procesamiento de lenguaje entre personas podría eventualmente comprender un lenguaje telegráfico, pero, al parecer, los modelos de procesamiento de lenguaje natural requieren estructuras sintácticas precisas u oraciones gramaticalmente bien formadas para procesar tareas específicas (Logan et al. 2021; Zamfirescu-Pereira et al. 2023; Leidinger et al. 2023). El número 9 hace una variación desestructurada del prompt 1, mientras que el número 10 hace una variación del prompt 2.

**Procesamiento de datos**

Luego de contar con la base de datos construida y los prompts diseñados, se instruyó a GPT-4o mini, a través de su API y del lenguaje Python, para que clasificara los 100.000 comentarios en positivos, negativos o neutrales con los diez *prompts*. El código utilizado fue construido a partir de la documentación de OpenAI para realizar análisis de sentimientos (Guzman 2024; OpenAI 2023a). Este código tiene varias secciones que pueden modificarse para que el procesamiento se ajuste. En este caso, dada la elección del código de análisis de sentimientos a partir de *batch*[5], los parámetros modificables fueron: modelo, temperatura y prompt, como se ve en la figura 1[6].

Se emplea el análisis de sentimientos como fuente principal en el código, ya que esta herramienta, ampliamente adoptada en los últimos años, facilita la clasificación de datos y permite una comparación más clara y práctica entre los resultados. No obstante, el objetivo de este artículo no es evaluar la eficacia del análisis de sentimientos, sino utilizarlo como un medio para reflexionar sobre el efecto de la estructuración de los prompts y el resultado del proceso de clasificación en sí mismo.

**Figura 1: código simplificado procesamiento de datos**

Sentiment analysis

The `sentiment_analysis` function analyzes the overall sentiment of the discussion. It considers the tone, the emotions conveyed by the language used, and the context in which words and phrases are used. For tasks which are less complicated, it may also be worthwhile to try out `gpt-3.5-turbo` in addition to `gpt-4` to see if you can get a similar level of performance. It might also be useful to experiment with taking the results of the `sentiment_analysis` function and passing it to the other functions to see how having the sentiment of the conversation impacts the other attributes.

```
1   def sentiment_analysis(transcription):
2       response = client.chat.completions.create(
3           model="gpt-4o mini",
4           temperature=0,
5           messages=[
6               {
7                   "role": "system",
8                   "content": "As an AI with expertise in language and emotion analysis, your task is to anal
9               },
10              {
11                  "role": "user",
12                  "content": transcription
13              }
14          ]
15      )
16      return completion.choices[0].message.content
```

**Fuente:** OpenAI. 2023d. *Sentiment analysis*. [Extracto]

---

[5] El procesamiento de datos se realizó utilizando el método batch de OpenAI, el cual permite enviar múltiples solicitudes de procesamiento en un formato JSON, optimizando tiempos y recursos. Este método implica dividir los datos en lotes (*batches*) y procesarlos en paralelo, manteniendo la consistencia de los resultados. Cada lote incluye parámetros como el modelo, la temperatura y los mensajes en formato JSON, tal y como se realiza generalmente. El flujo típico consiste en: (1) preparar los datos en lotes, (2) estructurarlos en JSON según la documentación de OpenAI, (3) enviarlos a la API utilizando un bucle, y (4) recopilar las respuestas para su análisis posterior.

[6] El código final tiene mayor complejidad por utilizar batch. Sin embargo, esta imagen simplifica los puntos centrales del código final.

Dado el objetivo de este artículo, la única variable manipulada fue el *prompt* con el objetivo de analizar los cambios en la clasificación de sentimientos que producen sus variaciones. La temperatura y el modelo se mantuvieron constantes. La temperatura del modelo se parametrizó en cero para mantener la estabilidad entre las respuestas (OpenAI 2023b). Los demás parámetros se dejaron en su versión predefinida y en todas las iteraciones se utilizó el modelo GPT-4o mini.

Con el código definido, se procesó la totalidad de la base de datos con cada uno de los *prompts* como instrucción. Esto se hizo de manera independiente para que el resultado del procesamiento de cada *prompt* no afectará el resultado del siguiente. Así, en cada petición a la API se enviaron solo los tokens de cada *prompt* y el texto a analizar de cada comentario. En total se trabajó con un millón de datos producto de procesar 100.000 datos diez veces.

**Aproximaciones cuantitativas**

Para testear la hipótesis de si cambios sutiles en la estructuración de *prompts* producen variaciones significativas en los resultados se recurrió a dos aproximaciones: exploratoria y confirmatoria. Por un lado, la exploratoria incluyó el Análisis de Componentes Principales (PCA) para analizar la similitud entre los *prompts* mediante su codificación en *embeddings*, junto con la matriz de coincidencias entre *prompts*, que refleja la proporción de clasificaciones equivalentes entre pares de comentarios. También se calculó la distancia de Levenshtein. Estas aproximaciones buscaban realizar un análisis preliminar sobre la clasificación de los *prompts* y su similitud, razón por la cual se consideran exploratorias.

Por otro lado, la aproximación confirmatoria se llevó a cabo mediante la prueba de Chi-cuadrado, con el objetivo de determinar si los *prompts* producían resultados significativamente diferentes entre sí. Este análisis buscaba responder a la pregunta de si las diferencias observadas en los resultados eran estadísticamente significativas.

**Análisis exploratorio**

Con respecto a la primera aproximación cuantitativa, se calcularon los *embeddings* de cada *prompt*. Los *embeddings* son representaciones vectoriales de alta dimensión generadas por modelos de lenguaje, que capturan el significado semántico de palabras o frases y permiten medir su similitud a través de la distancia en el espacio vectorial (OpenAI 2023c). Debido a la alta dimensionalidad de estos vectores, se aplicó un Análisis de Componentes Principales (PCA) para facilitar la visualización de los mismos. Esta metodología se seleccionó considerando que forma parte del conjunto de técnicas de reducción de dimensionalidad comúnmente utilizadas para la visualización de embeddings, como SVD y t-SNE (OpenAI 2022).

Para la segunda aproximación cuantitativa, se utilizó una matriz de coincidencia para comparar qué tanto se acercaban los *prompts* al clasificar los comentarios. Esta matriz muestra en una escala de 0 a 1 qué tanto coincide un par de *prompts*: entre más se acerque a 1 significa que la clasificación realizada por ambos fue más similar, mientras que una tendencia hacia 0 representa una mayor discrepancia.

Finalmente, la tercera aproximación fue la distancia de Levenshtein, también llamada distancia de edición, comúnmente usada para determinar la diferencia entre dos strings A y B, en términos de la cantidad de *caracteres* a agregar, quitar o sustituir al string A para obtener el string B o viceversa.

**Análisis confirmatorio**

En este procedimiento se utilizó una prueba Chi-cuadrado para evaluar si las funciones de densidad de probabilidad (PDF, del inglés *probability distribution function*) de los resultados de las clasificaciones (positivo, negativo, neutral) generadas por cada *prompt* tienen similitud. La prueba verifica si un *prompt* clasifica cada uno de los comentarios de manera similar a como lo hace otro *prompt*.

En este análisis, las frecuencias observadas ($O_i$) del evento $i$ representan el número de comentarios que fueron clasificados como positivos, negativos o neutrales por parte de uno de los *prompts*. Por un lado, las frecuencias ($E_i$) son los valores esperados de un evento $i$ usando otro *prompt* como referencia.

Este procedimiento asume que cada resultado de calificación solicitado por un *prompt* es independiente de otro, de esta forma la prueba Chi-cuadrado se puede hacer entre pares de resultados de los *prompts*.

Donde:

- $i$ representa los eventos de clasificación: positivo, negativo y neutral.

- N es el número total de comentarios clasificados, que en este caso fueron 100.000.

Estas frecuencias esperadas permitieron  calcular el estadístico de prueba que está dado por:

$$\chi^2 = \sum_{i=1}^{N} \frac{(O_i - E_i)^2}{E_i}$$

Este valor mide qué tan diferentes son las clasificaciones observadas del *prompt* en estudio respecto a las esperadas según el *prompt* de referencia y a partir de este valor se rechaza o no la hipótesis nula, a saber: las distribuciones de clasificaciones realizadas por el *prompt* en estudio no tienen diferencias estadísticamente significativas con respecto a las del *prompt* de referencia.

Con esto en consideración, se plantea la hipótesis alternativa: las distribuciones de clasificaciones realizadas por el *prompt* en estudio tienen diferencias estadísticamente significativas con respecto a las del *prompt* de referencia.

**Resultados**

Luego de procesar todos los comentarios con cada *prompt,* el resultado fue una tabla con los 100.000 comentarios y la manera en que cada *prompt* los clasificó. Así, en la tabla 2 se puede ver que un comentario como "Nunca vi a el karma de actuar de manera tan instantánea" (sic) fue clasificado como negativo por dos *prompts*, positivo por cinco *prompts* y neutral por tres *prompts*. Una minoría de los resultados del procesamiento del modelo fueron asignados a una cuarta categoría denominada inconsistente, si bien esta no hacía parte de las instrucciones dadas en el *prompt*. En este sentido, la categoría "inconsistente" reunió cualquier tipo de respuesta que no estuviese explícitamente relacionada con la clasificación de sentimientos del *prompt*: positivo, negativo o neutral.

**Tabla 2: Ejemplo de los resultados del proceso de clasificación**

| Texto Comentario | prompt 1 | prompt 2 | prompt 3 | prompt 4 | prompt 5 | prompt 6 | prompt 7 | prompt 8 | prompt 9 | prompt 10 |
|---|---|---|---|---|---|---|---|---|---|---|
| Que pedantes entrevistadores, ni siquiera pude acabar de ver la entrevista qué horror!!!!! Que no se dan cuenta con quién están hablando, una mujer muy inteligente brillante luz y esperanza para México!!!! Bendiciones para nuestra próxima presidenta!!!!! Porque este 2 de junio vamos a ganar, con Xóchilt | NEGATIVO | POSITIVO | NEGATIVO | POSITIVO | NEGATIVO | NEGATIVO | NEGATIVO | POSITIVO | NEGATIVO | NEGATIVO |
| Que factores de la Globalización determinaran el resultado y como se integra México a ese fenómeno sin verse afectado por las externalidades del neoliberalismo? | NEUTRAL | NEUTRAL | NEUTRAL | NEUTRAL | NEUTRAL | NEUTRAL | NEUTRAL | NEUTRAL | INCONSISTENTE | NEUTRAL |
| Que bueno !! Búkele será su talón de Aquiles | NEGATIVO | NEGATIVO | NEGATIVO | NEGATIVO | NEGATIVO | NEGATIVO | NEGATIVO | NEGATIVO | NEGATIVO | POSITIVO |
| Ay ahora si exige ya que le queda.a la botarga 😂😂😂😂😂 | NEGATIVO | NEGATIVO | NEGATIVO | NEGATIVO | NEGATIVO | NEGATIVO | NEGATIVO | NEGATIVO | NEGATIVO | POSITIVO |
| Nunca vi al karma de actuar de manera tan instantánea | NEGATIVO | POSITIVO | NEUTRAL | POSITIVO | NEUTRAL | POSITIVO | NEGATIVO | POSITIVO | NEUTRAL | POSITIVO |

**Fuente:** Elaboración propia

En la tabla 3, se observa el porcentaje de cada categoría para cada uno de los *prompts*. Esta tabla fue un primer acercamiento para preguntarse si los *prompts* clasificaron de la misma forma los comentarios. Como se puede observar, hay variaciones en estos porcentajes. Sin embargo, esta gráfica no permitió indagar las variaciones en la clasificación de un comentario en particular. Este método tampoco permitió verificar si las clasificaciones son estadísticamente diferentes, cosa que se comprobará en el análisis confirmatorio.

**Tabla 3: Porcentaje de la clasificación de sentimientos por *prompt***

| | NEGATIVO | POSITIVO | NEUTRAL | INCONSISTENTE |
|---|---|---|---|---|
| prompt 1 | 70.84 | 18.33 | 10.81 | 0.02 |
| prompt 2 | 66.87 | 22.14 | 10.98 | 0.00 |
| prompt 3 | 67.02 | 18.78 | 14.19 | 0.01 |
| prompt 4 | 66.85 | 22.62 | 10.53 | 0.00 |
| prompt 5 | 69.32 | 17.60 | 13.07 | 0.01 |
| prompt 6 | 66.33 | 21.12 | 12.54 | 0.00 |
| prompt 7 | 71.27 | 18.10 | 10.62 | 0.01 |
| prompt 8 | 65.89 | 22.50 | 11.61 | 0.00 |
| prompt 9 | 69.58 | 19.86 | 9.47 | 1.10 |
| prompt 10 | 66.48 | 23.20 | 10.31 | 0.01 |

**Fuente:** Elaboración propia

## Análisis de Componentes Principales

Con el propósito de comparar aquellos *prompts* que eran más parecidos entre sí (los cuales, en teoría, tendrían menos probabilidades de ser significativamente diferentes), se obtuvieron los *embeddings* de cada *prompt* con el objetivo de evaluar la similitud entre ellos.

**Figura 2. Similitud de *prompts* en el espacio de componentes principales**

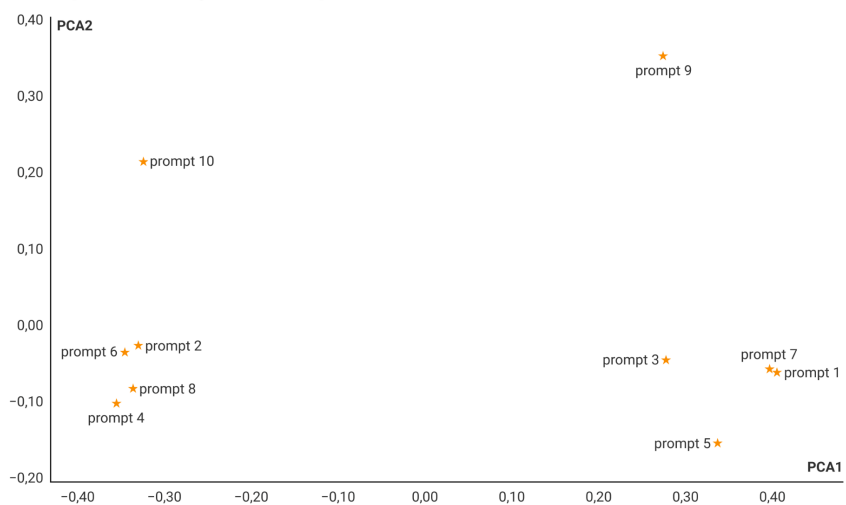

**Fuente:** Elaboración propia

Como se puede observar en la Figura 2, los *prompts* del grupo A tienden a agruparse en un lado del gráfico, mientras que los del B se ubican en el lado opuesto, con la excepción de los prompts 9 y 10, que presentan un comportamiento notablemente diferente al del resto. Esto puede deberse a que estos últimos fueron desestructurados lingüísticamente y se entienden como los más lejanos o, en este caso, diferentes.

Además, los *prompts* 1 y 7 del grupo A destacan por su cercanía en el cuadrante inferior derecho del gráfico. Esta proximidad podría indicar que el modelo interpreta ambas instrucciones de manera similar, probablemente debido a su parecido estructural. El prompt 1 es el de referencia sugerido por OpenAI y el 7 es una variación modal del 1. Su posición sugiere que estos *prompts* podrían generar resultados similares en la clasificación realizada por el modelo. Sin embargo, su contraposición en el grupo B, los *prompts* 2 y 8, que son el otro *prompt* de referencia y su variación modal respectivamente, no tienen tal cercanía.

Otros *prompts* con gran cercanía son los 2 y 6 del grupo B. El prompt 2 es el de referencia sugerido por Zhang et al. (2023) y el 6 es una variación léxico-semántica del 2. Nuevamente, su contraposición en el grupo A, los *prompts* 1 y 5 (el prompt de referencia de OpenAI y su variación léxico-semántica) no tienen tal cercanía. Lo anterior proporciona un marco inicial para conocer, contrastar y analizar los resultados de la prueba de Chi-cuadrado. Este análisis preliminar permite identificar patrones y posibles similitudes en la forma en que el modelo procesó cada uno de los *prompts*.

**Matriz de coincidencia**

Para cuantificar y hacer una aproximación inicial acerca de las diferencias y similitudes en la clasificación entre los *prompts*, se utilizó el cálculo de la proporción de coincidencias. Esta métrica, definida como el número de coincidencias exactas entre dos *prompts* dividido por el total de comparaciones realizadas, permitió evaluar de forma preliminar el nivel de concordancia entre los resultados obtenidos por diferentes *prompts* en la clasificación de sentimientos de los comentarios de los vídeos seleccionados.

**Tabla 4: Matriz de coincidencia**

|  | prompt 1 | prompt 2 | prompt 3 | prompt 4 | prompt 5 | prompt 6 | prompt 7 | prompt 8 | prompt 9 | prompt 10 |
|---|---|---|---|---|---|---|---|---|---|---|
| **prompt 1** | 1.00 | 0.93 | 0.95 | 0.93 | 0.96 | 0.93 | 0.98 | 0.93 | 0.94 | 0.92 |
| **prompt 2** | 0.93 | 1.00 | 0.94 | 0.97 | 0.93 | 0.96 | 0.93 | 0.97 | 0.92 | 0.96 |
| **prompt 3** | 0.95 | 0.94 | 1.00 | 0.94 | 0.96 | 0.94 | 0.95 | 0.94 | 0.92 | 0.93 |
| **prompt 4** | 0.93 | 0.97 | 0.94 | 1.00 | 0.92 | 0.96 | 0.93 | 0.97 | 0.92 | 0.96 |
| **prompt 5** | 0.96 | 0.93 | 0.96 | 0.92 | 1.00 | 0.93 | 0.96 | 0.92 | 0.93 | 0.92 |
| **prompt 6** | 0.93 | 0.96 | 0.94 | 0.96 | 0.93 | 1.00 | 0.92 | 0.96 | 0.91 | 0.94 |
| **prompt 7** | 0.98 | 0.93 | 0.95 | 0.93 | 0.96 | 0.92 | 1.00 | 0.92 | 0.94 | 0.92 |
| **prompt 8** | 0.93 | 0.97 | 0.94 | 0.97 | 0.92 | 0.96 | 0.92 | 1.00 | 0.92 | 0.96 |
| **prompt 9** | 0.94 | 0.92 | 0.92 | 0.92 | 0.93 | 0.91 | 0.94 | 0.92 | 1.00 | 0.92 |
| **prompt 10** | 0.92 | 0.96 | 0.93 | 0.96 | 0.92 | 0.94 | 0.92 | 0.96 | 0.92 | 1.00 |

**Fuente:** Elaboración propia

Como se evidencia en Tabla 4, los resultados mostraron coincidencias que oscilaron entre 0.92 y 0.98. Las diferencias son mayores cuando se comparan los *prompts* que eran basados en el oficial de OpenAI con los de Zhang et al. (2023) . Por ejemplo, si se compara el *prompt* 1 con el 2, los dos de base, hay una coincidencia de 0,93. Mientras que si se compara el 1 con el 7, hay una coincidencia de 0,98, porque el 7 era casi igual que el primero pero con la inclusión de verbos modales. La coincidencia de este par de *prompts* es consistente con las distancias en los *prompts* de la Figura 2.

Así mismo, la comparación entre el *prompt* 2 y el 6, con alta cercanía en la Figura 2, tiene un resultado de 0.96, que si bien es significativo no es tan alto como uno supondría por su cercanía. Este es el mismo valor que tienen el *prompt* 1 y 5, solo que estos en la Figura 2 no tenían tanta cercanía.

También se observa que los *prompts* desestructurados, sin palabras gramaticales ni puntuación, fueron los que mayor difirieron con el resto. Se puede sugerir la importancia de utilizar oraciones bien formadas y buena puntuación al momento de redactar *prompts*. Igualmente, se destaca que todos los *prompts* tuvieron variación: las únicas instancias en que el valor arrojó una coincidencia exacta de 1.0 se dieron cuando los *prompts* se compararon consigo mismos. El hecho de que las coincidencias variarán entre 0.92 y 0.98 plantea una reflexión alrededor de si estas diferencias son significativas y qué implicaciones pueden tener con respecto a la confianza y robustez de este modelo como se ha discutido en apartados anteriores.

**Distancia de Levenshtein**

Como complemento a lo anterior, se calculó la distancia de Levenshtein entre los tokens de cada *prompt* no a nivel de caracteres, sino de palabras. A cada palabra se le asoció un número único y dicha codificación se utilizó para convertir los *prompts* en arreglos numéricos. El algoritmo de Levenshtein se aplicó sobre dichos arreglos. En la práctica, los valores de la distancia de Levenshtein entre un *prompt* A y B son la cantidad de palabras que hay que agregar, quitar o sustituir para convertir el prompt A en el B o viceversa.

**Tabla 5. Matriz de distancias de Levenshtein**

|          | prompt 1 | prompt 2 | prompt 3 | prompt 4 | prompt 5 | prompt 6 | prompt 7 | prompt 8 | prompt 9 | prompt 10 |
|----------|----------|----------|----------|----------|----------|----------|----------|----------|----------|-----------|
| prompt 1 | 0 | 55 | 39 | 57 | 15 | 55 | 8 | 57 | 34 | 61 |
| prompt 2 | 55 | 0 | 63 | 13 | 54 | 1 | 56 | 6 | 32 | 17 |
| prompt 3 | 39 | 63 | 0 | 60 | 49 | 63 | 40 | 63 | 55 | 68 |
| prompt 4 | 57 | 13 | 60 | 0 | 58 | 14 | 58 | 17 | 34 | 23 |
| prompt 5 | 15 | 54 | 49 | 58 | 0 | 53 | 23 | 56 | 43 | 61 |
| prompt 6 | 55 | 1 | 63 | 14 | 53 | 0 | 56 | 7 | 32 | 18 |
| prompt 7 | 8 | 56 | 40 | 58 | 23 | 56 | 0 | 57 | 38 | 63 |
| prompt 8 | 57 | 6 | 63 | 17 | 56 | 7 | 57 | 0 | 33 | 23 |
| prompt 9 | 34 | 32 | 55 | 34 | 43 | 32 | 38 | 33 | 0 | 28 |
| prompt 10 | 61 | 17 | 68 | 23 | 61 | 18 | 63 | 23 | 28 | 0 |

**Fuente:** Elaboración propia

Como se evidencia en Tabla 5, la cantidad de palabras que hay que agregar, quitar o sustituir para convertir un *prompt* en el otro está entre 0 y 68. Las diferencias son mayores cuando se comparan los *prompts* basados en el oficial de OpenAI con los basados en Zhang et al. (2023). Por ejemplo, si se compara el *prompt* 1 con el 2, los dos de base, hay 55 palabras de distancia. En cambio, si se compara el 1 con el 7 hay una distancia de 8 palabras, porque el 7 era casi igual que el primero pero con la inclusión de verbos modales. La distancia de este par de *prompts* es consistente con las distancias en los prompts de la Figura 2 y la coincidencia de la Tabla 4.

Asimismo, la distancia de palabras entre el *prompt* 2 y el 6 es de tan solo 1 palabra, a pesar de que su coincidencia en la Tabla 4 es de 0,96. También, se observa que los *prompts* desestructurados, sin palabras gramaticales ni puntuación, no fueron los que mayor difirieron con el resto, como pasó en el análisis anterior. Finalmente, la correlación entre esta matriz y la matriz de coincidencia es de -0.71. Esto indica una fuerte correlación inversa: a menor distancia entre las palabras, mayor coincidencia.

Estas aproximaciones dan cuenta de forma preliminar de los resultados de clasificación de los *prompts* y sus diferencias, pero no determinan con exactitud si estas diferencias en la clasificación de resultados son significativas o no. Sin embargo, permitieron formular la hipótesis a testear en la prueba Chi-cuadrado.

**Prueba Chi-cuadrado**

Como se mencionó anteriormente, se decidió realizar la prueba Chi-cuadrado y calcular tanto el estadístico de prueba $\chi2$ como los *p-values* con un nivel de significancia del 5%. A partir de estos datos se determinó si los resultados de los *prompts* tienen la misma densidad de probabilidad (PDF) definida por Papoulis (2002).

Para el análisis cuantitativo, se transformaron los resultados categóricos: positivo, negativo, neutro e inconsistente en los siguientes valores numéricos: uno, dos, tres y cuatro respectivamente para el analisis Chi-cuadrado. De esta forma se creó un vector numérico por cada resultado de un *prompt* para hacer los cálculos.

Los *p-values* y el estadístico se encuentran consignados en las tablas 6 y 7 respectivamente.

**Tabla 6: p-values**

| | prompt 1 | prompt 2 | prompt 3 | prompt 4 | prompt 5 | prompt 6 | prompt 7 | prompt 8 | prompt 9 | prompt 10 |
|---|---|---|---|---|---|---|---|---|---|---|
| **prompt 1** | 1 | 9.31E-104 | 1.05E-124 | 6.66E-124 | 2.77E-54 | 5.82E-104 | **0.118** | 1.34E+138 | 2.10E-34 | 4.20E-157 |
| **prompt 2** | | 1 | 1.63E-149 | 7.26E-04 | 2.70E-162 | 6.14E-29 | 1.65E-120 | 9.65E-07 | 1.81E-62 | 6.79E-11 |
| **prompt 3** | | | 1 | 1.60E-195 | 1.61E-27 | 2.42E-53 | 2.57E-143 | 8.79E-132 | 1.57E-221 | 2.00E-235 |
| **prompt 4** | | | | 1 | 1.62E-205 | 2.38E-50 | 3.39E-140 | 1.04E-13 | 1.83E-62 | 4.64E-03 |
| **prompt 5** | | | | | 1 | 6.71E-87 | 5.16E-63 | 1.37E-168 | 2.17E-154 | 5.50E-251 |
| **prompt 6** | | | | | | 1 | 2.47E-124 | 2.10E-18 | 2.72E-118 | 3.47E-69 |
| **prompt 7** | | | | | | | 1 | 8.71E-160 | 5.04E-36 | 1.75E-174 |
| **prompt 8** | | | | | | | | 1 | 1.53E-104 | 2.92E-20 |
| **prompt 9** | | | | | | | | | 1 | 7.07E-79 |
| **prompt 10** | | | | | | | | | | 1 |

**Fuente:** Elaboración propia

**Tabla 7: estadísticos**

| | prompt 1 | prompt 2 | prompt 3 | prompt 4 | prompt 5 | prompt 6 | prompt 7 | prompt 8 | prompt 9 | prompt 10 |
|---|---|---|---|---|---|---|---|---|---|---|
| **prompt 1** | 0 | 474 | 571 | 567 | 247 | 475 | 4.27 | 635 | 155 | 720 |
| **prompt 2** | | 0 | 685 | 14.5 | 744 | 130 | 552 | 27.7 | 284 | 46.8 |
| **prompt 3** | | | 0 | 897 | 123 | 242 | 657 | 604 | 1020 | 1.08E+03 |
| **prompt 4** | | | | 0 | 943 | 229 | 642 | 60 | 284 | 10.7 |
| **prompt 5** | | | | | 0 | 397 | 287 | 773 | 708 | 1.15E+03 |
| **prompt 6** | | | | | | 0 | 569 | 81 | 541 | 315 |
| **prompt 7** | | | | | | | 0 | 732 | 163 | 800 |
| **prompt 8** | | | | | | | | 0 | 478 | 90 |
| **prompt 9** | | | | | | | | | 0 | 360 |
| **prompt 10** | | | | | | | | | | 0 |

**Fuente:** Elaboración propia

De las tablas 6 y 7 se concluye que solo los *prompts* 1 y 7 tienen la misma PDF, puesto que no se rechaza la hipótesis nula. En este sentido, solamente para este par de prompts se puede decir que el proceso de clasificación no tiene diferencias estadísticamente significativas. Esto está en consonancia con lo que sugiere tanto la Figura 2 como la Tabla 4.

Sin embargo, lo más interesante es que para el resto de los casos, la hipótesis nula es rechazada: los *p-values* son extremadamente bajos (menores que 0.05). Se puede concluir que para casi todos los pares de *prompts* se rechaza que no haya diferencias significativas en los resultados de clasificación para los mismos comentarios analizados.

Esto conlleva a rechazar la hipótesis del artículo y a afirmar que cambios sutiles en la estructuración de prompts, como por ejemplo de orden léxico, sintáctico, modal o su desestructuración, sí producen variaciones significativas en los resultados de clasificación del análisis de sentimientos de polaridad producido por el LLM GPT-4o mini. Estos hallazgos invitan a reflexionar sobre la confianza y la robustez en estos modelos, reflexión que daremos en el siguiente apartado de discusión.

**Discusión**

GPT-4o mini no siempre dio como respuesta los valores positivo, negativo o neutral. Los diez *prompts* lo instan a clasificar los comentarios únicamente en tres categorías pero, aún así, en todas las variaciones de la instrucción, al menos una respuesta fue inconsistente. En algunas ocasiones, el modelo mezcló categorías, incluyó nuevos sentimientos, brindó explicaciones e, incluso, utilizó otros idiomas diferentes al español, lo cual guarda relación con las variaciones léxicas, sintácticas y modales utilizadas en la construcción de los *prompts*. Todas estas respuestas no clasificadas en las categorías solicitadas fueron identificadas como *inconsistentes*. El *prompt* 9, el desestructurado del grupo A, fue el que más respuestas inconsistentes arrojó con más de 1000, seguido del *prompt* 10. Esto sugiere que la falta de oraciones bien formadas en los *prompts* y el uso de signos de puntuación puede generar una mayor cantidad de alucinaciones.

Otro hallazgo está en la robustez del LLM. El experimento muestra que, ante un análisis de sentimientos con solo tres opciones, el modelo no logró manejar las variaciones en los *prompts* sin afectar el proceso de clasificación. En este punto es posible introducir la pregunta por la confianza. ¿Cómo confiar en las clasificaciones que producen LLM con robustez limitada? Desde la perspectiva de Bolton et al. (2024), Huang et al. (2024) Koubaa et al. (2023), la confianza se cimenta en la robustez, pero, como anotamos, cambios menores en la forma de estructurar influyeron sensiblemente en la clasificación.

Esta limitación en la robustez es difícil de explicar. Como hemos mencionado, estos modelos funcionan como cajas negras inescrutables (Yadav y Vishwakarma 2020; Ermakova et al. 2021; Latour 2021). En este caso, GPT-4o mini nunca esclareció los criterios para definir si un comentario es positivo, negativo o neutral; tampoco si esos criterios-definiciones eran constantes o variables. Así las cosas, la confianza en el modelo no solo debería residir en la operatividad autocontenida que se supone robusta, pues como afirma Nowotny (2021) la incertidumbre es una característica inherente a los algoritmos predictivos.

Las reflexiones anteriores nos invitan a la siguiente pregunta: ¿podemos confiar en modelos inherentemente inciertos? Si la respuesta es afirmativa, es porque la confianza tiene que ver tanto con el cumplimiento de métricas o criterios preestablecidos (Bolton et al. 2024; Huang et al. 2024; Majeed y Hwang 2024) como con una relación social de atributo relacional (Cook y Santana 2020). La confianza en estos modelos puede darse a partir de que confiamos en la institución que los crea y mantiene (Cook, Hardin y Levi 2005), por experiencias sociales y tecnológicas previas, pero también por los incentivos asociados a ellos y a la dependencia que generan. Los incentivos tienen que ver con la capacidad que tienen para facilitar tareas diarias: "Trust as reliance in computer artifacts means that we expect an object to do something to help us attain our goals"(Goldberg 2020 and Coeckelbergh 2012 citados en Grodzinsky et al. 2020, 301). Así, la utilidad de estos modelos, y la confianza que generan, ha derivado en dependencia, o en términos de Smith (2024) en "obligatoriedad". Como propone Nowotny (2021), esa confianza es importante para dar una falsa sensación de certeza y control.

Estas perspectivas subrayan la relevancia de abordar tanto las potencialidades como los límites y los sesgos inherentes a los LLM (Páez 2021), temas que resultan cruciales en el contexto actual de su adopción creciente.

**Conclusiones y trabajo futuro**

La investigación evidencia que los LLM, como GPT-4o mini, son sensibles a variaciones sutiles en la estructuración de los prompts. Cambios léxicos, sintácticos, modales o su desestructuración generan cambios significativos en los resultados del análisis de sentimientos. Estas diferencias cuestionan la robustez del modelo, incluso cuando se utilizan instrucciones semánticamente similares.

Además, el modelo mostró respuestas inconsistentes, clasificando algunos comentarios fuera de las categorías solicitadas o proporcionando explicaciones no requeridas. Las instrucciones desestructuradas fueron particularmente propensas a generar alucinaciones. Esto sugiere que los LLM requieren *prompts* claros y oraciones gramaticalmente bien formadas para minimizar errores y mejorar la confiabilidad de los resultados.

En términos de confianza, se reflexiona que esta no solo se basa en el desempeño técnico de los modelos, sino en las relaciones sociales e institucionales que legitiman su uso. La dependencia creciente de estos modelos, combinada con la falsa percepción de certeza que generan, puede llevar a una aceptación acrítica de sus resultados. Este fenómeno resalta la necesidad de una adopción más reflexiva de los LLM en investigaciones y aplicaciones, sin descartar todas las potencialidades que tiene para analizar fenómenos sociales.

Para trabajos futuros el camino experimental continúa. Un próximo paso clave será comparar los resultados obtenidos por los diferentes prompts con un *ground truth*, es decir, una clasificación realizada por humanos, para identificar cuáles se acercan más a las evaluaciones humanas. Además, resulta fundamental probar la hipótesis planteada en este estudio en otros idiomas, con énfasis en el inglés, para evaluar la consistencia del modelo en contextos lingüísticos diversos y así ampliar la validez de los hallazgos.

## Referencias

**Anexo 1. Canales periodísticos de los que se recolectó la información**

## YouTube News Channels Information

| News Channel Name | Category | Link | Channel ID |
|---|---|---|---|
| France 24 Español | Internacional | https://www.youtube.com/@France24_es | UCUdOoVWuWmgo1wByzcsyKDQ |
| Euronews en Español | Internacional | https://www.youtube.com/@euronewses | UCyoGb3SMlTlB8CLGVH4c8Rw |
| CNN en Español | Internacional | https://www.youtube.com/@cnnee | UCJag7ggH_TnSye399htVAEw |
| DW Español | Internacional | https://www.youtube.com/@dwespanol | UChmI8e6G2mV-0zf9SuQdbVA |
| Telemundo | Internacional | https://www.youtube.com/@noticias | UC2Xq2PK-got3Rtz9ZJ32hLQ |
| TN - Todo Noticias | Argentina | https://www.youtube.com/@todonoticias | UCyS7HkalUX2FBJkJUsnFZUA |
| C5N | Argentina | https://www.youtube.com/@c5n | UCFgk2Q2mVO1BklRQhSv6p0w |
| Canal 26 | Argentina | https://www.youtube.com/@canal26 | UCrpMfcQNog595v5gAS-oUsQ |
| La Nación | Argentina | https://www.youtube.com/@lanacion | UCba3hpU7EFBSk817y9qZkiA |
| A24 | Argentina | https://www.youtube.com/@A24com | UCR9120YBAqMfntqgRTKmkjQ |
| Noticias Caracol | Colombia | https://www.youtube.com/@noticiascaracol | UC2Xq2PK-got3Rtz9ZJ32hLQ |
| Noticias RCN | Colombia | https://www.youtube.com/@NoticiasRCN | UCBlaAnzlF0yOwhTJxXOKluA |
| El Tiempo | Colombia | https://www.youtube.com/@ElTiempo | UCe5-b0fCK3eQCpwS6MT0aNw |
| El espectador | Colombia | https://www.youtube.com/@ElEspectador | UCFYdMS-M4mRdbxPuErz1CVw |
| Noticas Uno | Colombia | https://www.youtube.com/@NoticiasUnoColombia | UC3NUgxBBl05tthfGBS6kPkA |
| Nmás | México | https://www.youtube.com/@nmas | UCUsm-fannqOY02PNN67C0KA |
| TV Azteca Noticias | México | https://www.youtube.com/@aztecanoticias | UCUP6qv-_EIL0hwTsJaKYnvw |
| Milenio Televisión | México | https://www.youtube.com/@milenio | UCFxHplbcoJK9m70c4VyTIxg |
| Imagen Televisión | México | https://www.youtube.com/ImagenNoticias | UCcHtke0raf8vNYLM1Hbq3Jw |
| Excélsior TV | México | https://www.youtube.com/@ExcelsiorMex | UClqo4ZAAZ01HQdCTIovCgkA |
| La Prensa Gráfica Noticias de El Salvador | El Salvador | https://www.youtube.com/@laprensagraficavideo | UCNz2y0qMguce8wyxl0oFXUw |
| Canal 12 Noticias | El Salvador | https://www.youtube.com/@Canal12ElSalvador | UC4rvyYYuqGGti3-u6VjN4SA |
| elsalvadorcom | El Salvador | https://www.youtube.com/c/elsalvadorcomyoutube | UCr865ABWoDU_qMeQ5LUrbyw |
| Factum | El Salvador | https://www.youtube.com/@revistafactumsv | UC8tvT2mf0LuoY-y_a8UAlpQ |

Elaboración propia